# Kalman's shrinkage for wavelet-based despeckling of SAR images.

Mario Mastriani, and Alberto E. Giraldez

***Abstract*—In this paper, a new probability density function (pdf) is proposed to model the statistics of wavelet coefficients, and a simple Kalman's filter is derived from the new pdf using Bayesian estimation theory. Specifically, we decompose the speckled image into wavelet subbands, we apply the Kalman's filter to the high subbands, and reconstruct a despeckled image from the modified detail coefficients. Experimental results demonstrate that our method compares favorably to several other despeckling methods on test synthetic aperture radar (SAR) images.

## I. Introduction

THE despeckling of a SAR image corrupted by speckle noise is an important problem in image processing. If the wavelet transform and maximum a-posteriori (MAP) estimator are used for this problem, the solution requires a priori knowledge about wavelet coefficients. Therefore, two problems arise: 1) What kind of distribution represents the wavelet coefficients ? 2) What is the corresponding estimator (shrinkage function) ?

Statistical models pretend wavelet coefficients to be random realizations from a distribution function. For the first problem, mostly these models assume the coefficients are time-invariant, and try to characterize them by using Gaussian, Laplacian, or Gaussian scale mixtures. For example, classical soft thresholding operator suggested by Donoho [1] can be obtained by a Laplacian assumption. Bayesian methods for image denosing using other distributions have also been proposed [2]–[6].

The wavelet transform has become an important tool for this problem due to its energy compaction property. Basically, it states that the wavelet transform yields a large number of small coefficients and a small number of large coefficients.

Simple denoising algorithms that use the wavelet transform consist of three steps:

1) Calculate the wavelet transform of the noisy image.
2) Modify the noisy wavelet coefficients according to some rule.
3) Compute the inverse transform using the modified coefficients.

Besides, two properties are very important for the following deduction: 1) If a wavelet coefficient is large/small, the adjacent coefficients are likely to be large/small, and 2) large/small coefficients tend to propagate across the scales [7]-[9]. In this paper, we use a new time-variant pdf and derive a Kalmanian shrinkage (KalmanShrink) function using Bayesian estimation theory.

## II. Bayesian Denoising

In this section, the denoising of an image corrupted by white Gaussian noise will be considered, i.e.,

$$g = x + n \qquad (1)$$

where $n$ is independent Gaussian noise. We observe $g$ (a noisy signal) and wish to estimate the desired signal $x$ as accurately as possible according to some criteria. In the wavelet domain, if we use an orthogonal wavelet transform, the problem can be formulated as

$$y = w + n \qquad (2)$$

where $y$ noisy wavelet coefficient, $w$ true coefficient, and $n$ noise, which is independent Gaussian. This is a classical problem in estimation theory. Our aim is to estimate from the noisy observation. The MAP estimator will be used for this purpose. Time-variant and time-invariant models will be discussed for this problem in Sections II-A and B, and new MAP estimators are derived.

### A. Time-Invariant Models

The classical MAP estimator for (2) is

$$\hat{w}(y) = \arg\max_{w} p_{w/y}(w/y). \qquad (3)$$

Using Bayes rule, one gets

$$\hat{w}(y) = \arg\max_{w} [p_{y/w}(y/w) \cdot p_w(w)]$$

$$= \arg\max_{w} [p_n(y-w) \cdot p_w(w)]. \qquad (4)$$

Therefore, these equations allow us to write this estimation in terms of the pdf of the noise ($p_n$) and the pdf of the signal coefficient ($p_w$). From the assumption on the noise, $p_n$ is zero mean Gaussian with variance $\sigma_n$, i.e.,

$$p_n(n) = \frac{1}{\sigma_n \sqrt{2\pi}} \cdot \exp\left(-\frac{n^2}{2\sigma_n^2}\right). \quad (5)$$

It has been observed that wavelet coefficients of natural images have highly non-Gaussian statistics [7]-[9]. The pdf for wavelet coefficients is often modeled as a generalized (heavy-tailed) Gaussian [7]-[9].

$$p_w(w) = K(s,p) \cdot \exp\left(-\left|\frac{w}{s}\right|^p\right). \quad (6)$$

where $s$, $p$ are the parameters for this model, and $K(s,p)$ is the parameter-dependent normalization constant. Other pdf models have also been proposed [7]-[9]. In practice, generally, two problems arise with the Bayesian approach when an accurate but complicated pdf $p_w(w)$ is used: 1) It can be difficult to estimate the parameters of $p_w$ for a specific image, especially from noisy data, and 2) the estimators for these models may not have simple closed form solution and can be difficult to obtain. The solution for these problems usually requires numerical techniques.

Let us continue developing the MAP estimator and show it for Gaussian and Laplacian cases. Equation (4) is also equivalent to

$$\hat{w}(y) = \arg\max_w \left[\log(p_n(y-w)) + \log(p_w(w))\right]. \quad (7)$$

As in [7]-[9], let us define $f(w) = \log(p_w(w))$. By using (5), (7) becomes

$$\hat{w}(y) = \arg\max_w \left[-\frac{(y-w)^2}{2\sigma_n^2} + f(w)\right]. \quad (8)$$

This is equivalent to solving the following equation for $\hat{w}$ if $p_w(w)$ is assumed to be strictly convex and differentiable.

$$\frac{y - \hat{w}}{\sigma_n^2} + f'(\hat{w}) = 0. \quad (9)$$

If $p_w(w)$ is assumed to be a zero mean gaussian density with variance $\sigma^2$, then $f(w) = -\log(\sqrt{2\pi}\sigma) - w^2/2\sigma^2$, and the estimator can be written as

$$\hat{w}(y) = \frac{\sigma^2}{\sigma^2 + \sigma_n^2} \cdot y. \quad (10)$$

If it is Laplacian

$$p_w(w) = \frac{1}{\sqrt{2}\sigma} \exp\left(-\frac{\sqrt{2}|w|}{\sigma}\right). \quad (11)$$

then $f(w) = -\log(\sigma\sqrt{2}) - \sqrt{2}|w|/\sigma$, and the estimator will be

$$\hat{w}(y) = \text{sign}(y)\left(|y| - \frac{\sqrt{2}\sigma_n^2}{\sigma}\right)_+. \quad (12)$$

Here, $(g)_+$ is defined as

$$(g)_+ = \begin{cases} 0, & \text{if } g < 0 \\ g, & \text{otherwise} \end{cases} \quad (13)$$

Equation (12) is the classical soft shrinkage function. Let us define the soft operator as

$$\text{soft}(g, \tau) = \text{sign}(g) \cdot (|g| - \tau)_+. \quad (14)$$

The soft shrinkage function (12) can be written as

$$\hat{w}(y) = \text{soft}\left(y, \frac{\sqrt{2}\sigma_n^2}{\sigma}\right). \quad (15)$$

*B. Time-Variant Model (KalmanShrink)*

Time-invariant models cannot model the temporal variation of wavelet coefficients. However, the coefficients depend strongly on the time. This paper suggests such dependency and derives the corresponding Kalmanian MAP estimator based on noisy wavelet coefficients in detail.

Here, we modify the Bayesian estimation problem as to take into account the time dependency of coefficients. If pdf is

$$p_w(w) = \exp\left(-\frac{1}{\sigma_n^2 K_t} \sum_w (\hat{w}_{t+1}(y) - \hat{w}_t(y))\right). \quad (16)$$

then

$$f(\hat{w}) = -\frac{1}{\sigma_n^2 K_t} \sum_w (\hat{w}_{t+1}(y) - \hat{w}_t(y)). \quad (17)$$

with

$$f'(\hat{w}) = -\frac{(\hat{w}_{t+1}(y) - \hat{w}_t(y))}{\sigma_n^2 K_t}. \quad (18)$$

Replacing (18) into (9), the estimator can be written as

$$\hat{w}_{t+1}(y) = \hat{w}_t(y) + K_t.(y_t - \hat{w}_t). \qquad (19)$$

and

$$\hat{y}_t = \hat{w}_t(y). \qquad (20)$$

where $K_t$ is the Kalman's gain of the Kalman's filter [10]-[15], being

$$K_t = \frac{P_t}{P_t + \sigma_n^2} \qquad (21)$$

and

$$P_t = (1 - K_t).P_t \qquad (22)$$

is the norm of error covariance, that is to say

$$P_t = \left\| E\{(y_t - \hat{w}_t).(y_t - \hat{w}_t)^T\} \right\|_2. \qquad (23)$$

where $[\bullet]^T$ means transpose of $[\bullet]$, $E\{\bullet\}$ means the expectation of $\{\bullet\}$ and $\|(\bullet)\|_2$ means the $L_2$-norm of $(\bullet)$. Finally, the original system is formed by (2) and

$$w_{t+1} = w_t. \qquad (24)$$

These equations can be described as a Kalman filter loop [10] as follows:

1) Calculate $\sigma_n^2$ as the variance of (2)
2) Initialize $\hat{w}$ and $P$
3) Choose $P_{final}$ as a finish condition
4) *while* $P_t > P_{final}$,

   Compute Kalman's gain $K_t$ from (21)

   Update estimate $\hat{w}_{t+1}(y)$ with measurement $y_t$ from (19)

   Compute error covariance $P_t$ for updated estimate from (22)
   *end while*.

### III. SPECKLE MODEL

Speckle noise in SAR images is usually modeled as a purely multiplicative noise process of the form

$$I_s(r,c) = I(r,c).S(r,c)$$

$$= I(r,c).[1 + S'(r,c)]$$

$$= I(r,c) + N(r,c) \qquad (25)$$

The true radiometric values of the image are represented by $I$, and the values measured by the radar instrument are represented by $I_s$. The speckle noise is represented by $S$. The parameters $r$ and $c$ means row and column of the respective pixel of the image. If

$$S'(r,c) = S(r,c) - 1 \qquad (26)$$

and

$$N(r,c) = I(r,c).S'(r,c) \qquad (27)$$

we begin with a multiplicative speckle $S$ and finish with an additive speckle $N$ [16], which avoid the log-transform, because the mean of log-transformed speckle noise does not equal to zero [17] and thus requires correction to avoid extra distortion in the restored image.

For single-look SAR images, $S$ is Rayleigh distributed (for amplitude images) or negative exponentially distributed (for intensity images) with a mean of *1*. For multi-look SAR images with independent looks, $S$ has a gamma distribution with a mean of *1*. Further details on this noise model are given in [18].

### IV. ASSESSMENT PARAMETERS

In this work, the assessment parameters that are used to evaluate the performance of speckle reduction are Noise Variance, Mean Square Difference, Noise Mean Value, Noise Standard Deviation, Equivalent Number of Looks, Deflection Ratio [19], [20], and Pratt's figure of Merit [21].

*A. Noise Mean Value (NMV), Noise Variance (NV), and Noise Standard Deviation (NSD)*

*NV* determines the contents of the speckle in the image. A lower variance gives a "cleaner" image as more speckle is reduced, although, it not necessarily depends on the intensity. The formulas for the *NMV*, *NV* and *NSD* calculation are

$$NMV = \frac{\sum_{r,c} I_d(r,c)}{R*C} \qquad (28.1)$$

$$NV = \frac{\sum_{r,c} (I_d(r,c) - NMV)^2}{R*C} \qquad (28.2)$$

$$NSD = \sqrt{NV} \qquad (28.3)$$

where *R-by-C* pixels is the size of the despeckled image $I_d$. On the other hand, the estimated noise variance is used to determine the amount of smoothing needed for each case for all filters.

*B. Mean Square Difference (MSD)*

*MSD* indicates average square difference of the pixels throughout the image between the original image (with speckle) $I_s$ and $I_d$, see Fig. 1. A lower *MSD* indicates a smaller difference between the original (with speckle) and despeckled image. This means that there is a significant filter performance. Nevertheless, it is necessary to be very careful with the edges. The formula for the *MSD* calculation is

$$MSD = \frac{\sum_{r,c}(I_s(r,c) - I_d(r,c))^2}{R*C} \quad (29)$$

*C. Equivalent Numbers of Looks (ENL)*

Another good approach of estimating the speckle noise level in a SAR image is to measure the *ENL* over a uniform image region [20]. A larger value of *ENL* usually corresponds to a better quantitative performance. The value of *ENL* also depends on the size of the tested region, theoretically a larger region will produces a higher *ENL* value than over a smaller region but it also tradeoff the accuracy of the readings. Due to the difficulty in identifying uniform areas in the image, we proposed to divide the image into smaller areas of 25x25 pixels, obtain the *ENL* for each of these smaller areas and finally take the average of these *ENL* values. The formula for the *ENL* calculation is

$$ENL = \frac{NMV^2}{NSD^2} \quad (30)$$

The significance of obtaining both *MSD* and *ENL* measurements in this work is to analyze the performance of the filter on the overall region as well as in smaller uniform regions.

*D. Deflection Ratio (DR)*

A fourth performance estimator that we used in this work is the *DR* proposed by H. Guo et al (1994), [22]. The formula for the deflection calculation is

$$DR = \frac{1}{R*C}\sum_{r,c}\left(\frac{I_d(r,c) - NMV}{NSD}\right) \quad (31)$$

The ratio *DR* should be higher at pixels with stronger reflector points and lower elsewhere. In H. Guo *et al*'s paper, this ratio is used to measure the performance between different wavelet shrinkage techniques. In this paper, we apply the ratio approach to all techniques after despeckling in the same way [19].

*E. Pratt's figure of merit (FOM)*

To compare edge preservation performances of different speckle reduction schemes, we adopt the Pratt's figure of merit [21] defined by

$$FOM = \frac{1}{max\{\hat{N}, N_{ideal}\}} \sum_{i=1}^{\hat{N}} \frac{1}{1+d_i^2 \alpha} \quad (32)$$

Where $\hat{N}$ and $N_{ideal}$ are the number of detected and ideal edge pixels, respectively, $d_i$ is the Euclidean distance between the *i*th detected edge pixel and the nearest ideal edge pixel, and α is a constant typically set to 1/9. *FOM* ranges between *0* and *1*, with unity for ideal edge detection.

## V. EXPERIMENTAL RESULTS

Here, we present a set of experimental results using one ERS SAR Precision Image (PRI) standard of Buenos Aires area. For statistical filters employed along, i.e., Median, Lee, Kuan, Gamma-Map, Enhanced Lee, Frost, Enhanced Frost [19], [20], Wiener [23], DS [21] and Enhanced DS (EDS) [19], we use a homomorphic speckle reduction scheme [19], with 3-by-3, 5-by-5 and 7-by-7 kernel windows. Besides, for Lee, Enhanced Lee, Kuan, Gamma-Map, Frost and Enhanced Frost filters the damping factor is set to 1 [19], [20].

Fig. 1 shows a noisy image used in the experiment from remote sensing satellite ERS-2, with a 242-by-242 (pixels) by 65536 (gray levels); and the filtered images, processed by using VisuShrink (Hard-Thresholding), BayesShrink, NormalShrink, SUREShrink, and KalmanShrink techniques respectively, see Table I. All the wavelet-based techniques used Daubechies 1 wavelet basis and 1 level of decomposition (improvements were not noticed with other basis of wavelets) [4], [21], [23]. Besides, Fig. 1 summarizes the edge preservation performance of the KalmanShrink technique vs. the rest of the shrinkage techniques with a considerably acceptable computational complexity.

Table I shows the assessment parameters vs. 19 filters for Fig. 1, where En-Lee means Enhanced Lee Filter, En-Frost means Enhanced Frost Filter, Non-log SWT means Non-logarithmic Stationary Wavelet Transform Shrinkage [16], Non-log DWT means Non-logarithmic DWT Shrinkage [17], VisuShrink (HT) means Hard-Thresholding, (ST) means Soft-Thresholding, and (SST) means Semi-ST [4], [20], [22]-[28].

We compute and compare the NMV and NSD over six different homogeneous regions in our SAR image, before and after filtering, for all filters. The KalmanAShrink has obtained the best mean preservation and variance reduction, as shown in Table I. Since a successful speckle reducing filter will not significantly affect the mean intensity within a homogeneous region, KalmanShrink demonstrated to be the best in this sense too. The quantitative results of Table I show that the KalmanShrink technique can eliminate speckle without distorting useful image information and without destroying the important image edges. In fact, the KalmanShrink outperformed the conventional and no conventional speckle reducing filters in terms of edge preservation measured by

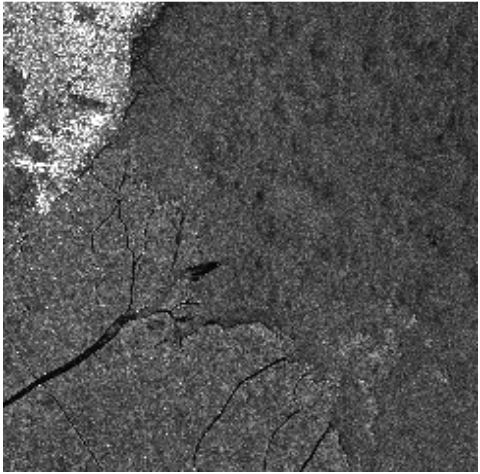
(a) original

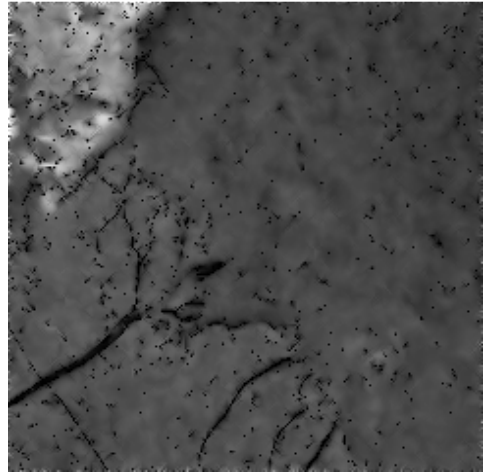
(b) VisuShrink

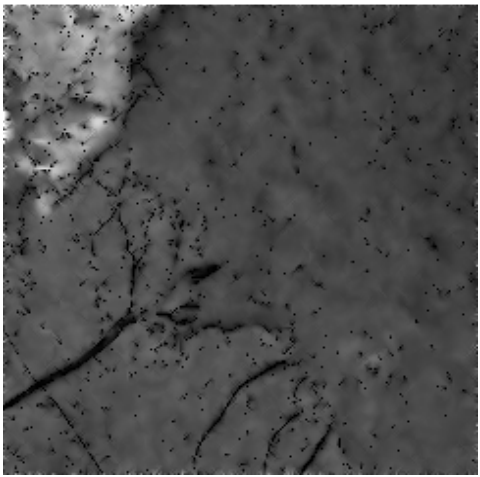
(c) BayesShrink

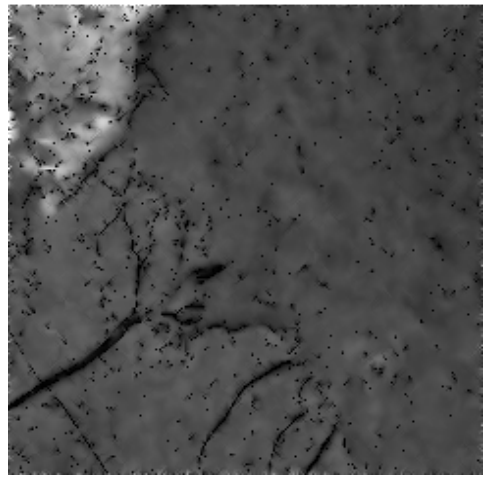
(d) NormalShrink

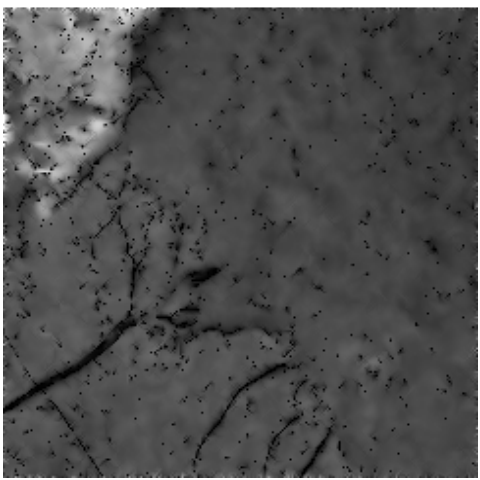
(e) SUREShrink

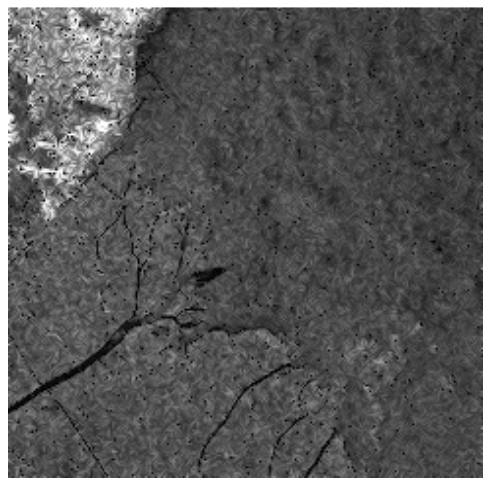
(f) KalmanShrink

Fig. 1: Original and filtered images.

Table I. Assessment Parameters vs. Filters for Fig. 1.

| Filter | Assessment Parameters | | | | | |
|---|---|---|---|---|---|---|
| | MSD | NMV | NSD | ENL | DR | FOM |
| Original noisy image | - | 90.0890 | 43.9961 | 11.0934 | 2.5580e-017 | 0.3027 |
| En-Frost | 564.8346 | 87.3245 | 40.0094 | 16.3454 | 4.8543e-017 | 0.4213 |
| En-Lee | 532.0006 | 87.7465 | 40.4231 | 16.8675 | 4.4236e-017 | 0.4112 |
| Frost | 543.9347 | 87.6463 | 40.8645 | 16.5331 | 3.8645e-017 | 0.4213 |
| Lee | 585.8373 | 87.8474 | 40.7465 | 16.8465 | 3.8354e-017 | 0.4228 |
| Gamma-MAP | 532.9236 | 87.8444 | 40.6453 | 16.7346 | 3.9243e-017 | 0.4312 |
| Kuan | 542.7342 | 87.8221 | 40.8363 | 16.9623 | 3.2675e-017 | 0.4217 |
| Median | 614.7464 | 85.0890 | 42.5373 | 16.7464 | 2.5676e-017 | 0.4004 |
| Wiener | 564.8346 | 89.8475 | 40.3744 | 16.5252 | 3.2345e-017 | 0.4423 |
| DS | 564.8346 | 89.5353 | 40.0094 | 17.8378 | 8.5942e-017 | 0.4572 |
| EDS | 564.8346 | 89.3232 | 40.0094 | 17.4242 | 8.9868e-017 | 0.4573 |
| VisuShrink (HT) | 855.3030 | 88.4311 | 32.8688 | 39.0884 | 7.8610e-016 | 0.4519 |
| VisuShrink (ST) | 798.4422 | 88.7546 | 32.9812 | 38.9843 | 7.7354e-016 | 0.4522 |
| VisuShrink (SST) | 743.9543 | 88.4643 | 32.9991 | 37.9090 | 7.2653e-016 | 0.4521 |
| SureShrink | 716.6344 | 87.9920 | 32.8978 | 38.3025 | 2.4005e-015 | 0.4520 |
| NormalShrink | 732.2345 | 88.5233 | 33.3124 | 36.8464 | 6.7354e-016 | 0.4576 |
| BayesShrink | 724.0867 | 88.9992 | 36.8230 | 36.0987 | 1.0534e-015 | 0.4581 |
| Non-log SWT | 300.2841 | 86.3232 | 43.8271 | 11.2285 | 1.5783e-016 | 0.4577 |
| Non-log DWT | 341.3989 | 87.1112 | 39.4162 | 16.4850 | 1.0319e-015 | 0.4588 |
| KalmanShrink | 867.1277 | 90.0890 | 32.6884 | 39.0884 | 3.2675e-015 | 0.4591 |

Pratt's figure of merit [21], as shown in Table I.

On the other hand, all filters were implemented in MATLAB® (Mathworks, Natick, MA) on a PC with an Athlon (2.4 GHz) processor.

## VI. CONCLUSION

We have presented a new speckle filter for SAR images based on wavelet denoising. In order to convert the multiplicative speckle model into an additive noise model, Argenti *et al*'s approach is applied. The simulations show that the KalmanShrink have better performance than the most commonly used filters for SAR imagery (for the studied benchmark parameters) which include statistical filters and several wavelets techniques in terms of smoothing uniform regions and preserving edges and features. The effectiveness of the technique encourages the possibility of using the approach in a number of ultrasound and radar applications. In fact, cleaner images suggest potential improvements for classification and recognition. Besides, considerably increased deflection ratio strongly indicates improvement in detection performance.

Finally, the method is computationally efficient and can significantly reduce the speckle while preserving the resolution of the original image, and avoiding several levels of decomposition and block effect.

ACKNOWLEDGMENT

M. Mastriani thanks Prof. Horacio Franco from Stanford Research Institute (SRI) for his help and support.